\documentclass[10pt,twocolumn,letterpaper]{article}

\usepackage[pagenumbers]{cvpr} 

\usepackage{graphicx}
\usepackage{amsmath}
\usepackage{amssymb}
\usepackage{booktabs}
\usepackage{times}
\usepackage{epsfig}
\usepackage{float}
\usepackage{multirow}
\usepackage{enumitem}

\usepackage{algorithm}
\usepackage{algpseudocode}
\usepackage{graphicx, amsmath, amssymb, caption, subcaption, multirow, overpic, textpos}
\usepackage[table,dvipsnames]{xcolor}
\usepackage[british, english, american]{babel}
\definecolor{citecolor}{HTML}{0071BC}
\definecolor{linkcolor}{HTML}{ED1C24}

\usepackage{color}
\usepackage{xcolor}
\definecolor{citecolor}{HTML}{0071bc}
\usepackage[pagebackref=true,breaklinks=true,colorlinks,citecolor=citecolor,bookmarks=false]{hyperref}

\usepackage[capitalize]{cleveref}
\crefname{section}{Sec.}{Secs.}
\Crefname{section}{Section}{Sections}
\Crefname{table}{Table}{Tables}
\crefname{table}{Tab.}{Tabs.}

\def\confName{CVPR}
\def\confYear{2023}

\begin{document}

\title{Policy Adaptation from Foundation Model Feedback}

	\author{Yuying Ge$^{1\star}$ \quad
		Annabella Macaluso$^2$ \quad 
		Li Erran Li$^{3\dagger}$ \quad
		Ping Luo$^1$ \quad 
		Xiaolong Wang$^2$\\
		{$^1$University of Hong Kong} \quad {$^2$University of California, San Diego} \quad {$^3$AWS AI, Amazon}\\
  {\small \url{https://geyuying.github.io/PAFF}}
	}
 \renewcommand{\thefootnote}{\fnsymbol{footnote}}
 		\footnotetext[1]{Work done during internship at UCSD.} 
   \footnotetext[2]{Work done outside of Amazon.} 
		\begin{figure}[htb]
		\twocolumn[{
			\maketitle
			\vspace{-20pt}
			\begin{center}
				\centering
				\vspace{-6pt}
				\includegraphics[width=1.0\textwidth]{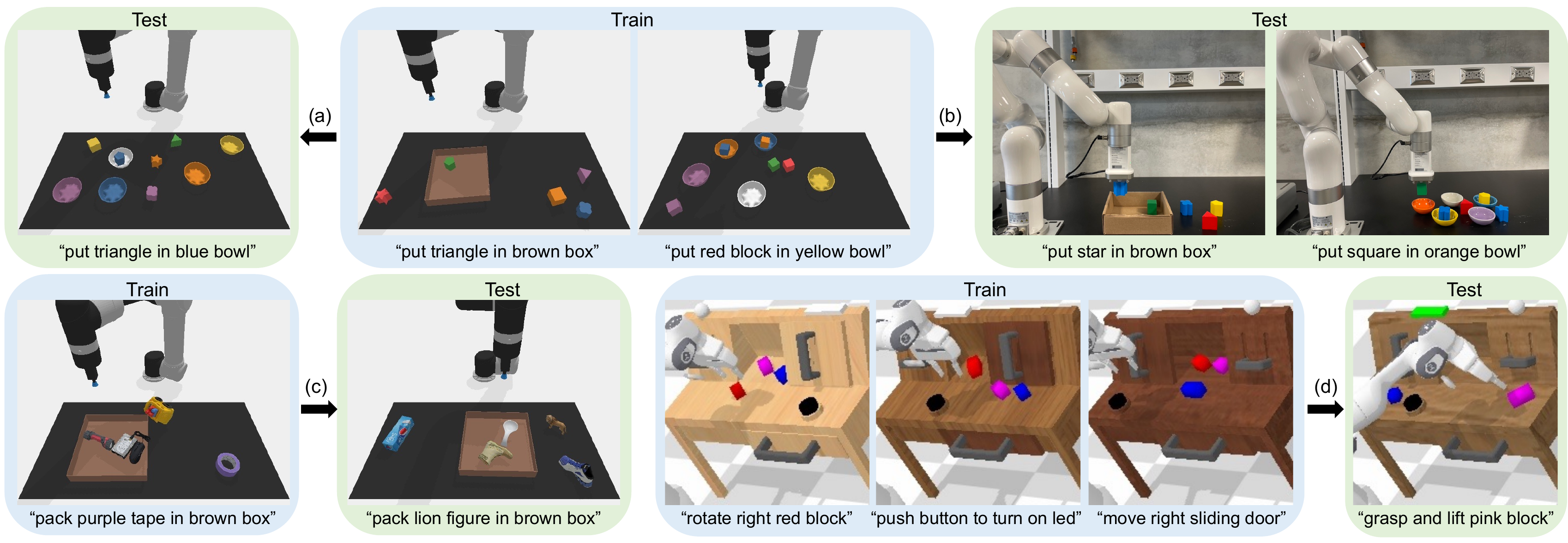}
				\vspace{-9mm}
			\end{center}
			\caption{
   Examples of the language conditioned policy adaptation experiments, including the evaluation of (i) \textbf{Compositional Generalization} in (a), where we train a policy to pack \textit{objects of different shapes} in the brown box, and put blocks of different colors in the \textit{bowls of different colors}, and adapt it to put \textit{objects of different shapes} in the \textit{bowls of different colors}; (ii) \textbf{Out-of-distribution Generalization} in (c), where we train a policy to pack certain objects in the brown box, and adapt it to pack \textit{unseen objects}, and in (d), where we adapt a policy trained on seen environments to \textit{an unseen environment} with different textures and placements of static elements such as the sliding door and the light button; (iii) \textbf{Sim-to-real Transfer} in (b), where we adapt a policy trained on simulation data to the \textit{real world}.}
			\label{fig:tasks}
			\vspace{15pt}
		}	]
	\end{figure}

\vspace{-0.2in}
\begin{abstract}
\vspace{-0.05in}
Recent progress on vision-language foundation models have brought significant advancement to building general-purpose robots. By using the pre-trained models to encode the scene and instructions as inputs for decision making, the instruction-conditioned policy can generalize across different objects and tasks. While this is encouraging, the policy still fails in most cases given an unseen task or environment. In this work, we propose Policy Adaptation from Foundation model Feedback (PAFF). When deploying the trained policy to a new task or a new environment, we first let the policy play with randomly generated instructions to record the demonstrations. While the execution could be wrong, we can use the pre-trained foundation models to provide feedback to relabel the demonstrations. This automatically provides new pairs of demonstration-instruction data for policy fine-tuning. We evaluate our method on a broad range of experiments with the focus on generalization on unseen objects, unseen tasks, unseen environments, and sim-to-real transfer. We show PAFF improves baselines by a large margin in all cases. 
\end{abstract}

\section{Introduction}
\label{sec:intro}
Learning generalizable manipulation policies have been a long standing problem in robotics. The goal is to train a general-purpose robot which can tackle multiple tasks with different object compositions in diverse environments. However, most current policy learning approaches with imitation learning or reinforcement learning can only learn to solve one task at a time, and usually operate on a fixed set of objects. To achieve human-level generalization, many efforts have been made on performing robotic manipulation tasks specified by natural language~\cite{cliport, hulc, calvin}. Language can not only bring its compositionality to low-level robot skills, but also operate as a high-level planner for long-horizon tasks.

Recently, there is a trend on leveraging the pre-trained vision-language foundation models~\cite{clip,slip} as the backbone encoders for generalizing robot skills. For example, CLIPORT~\cite{cliport} uses the pre-trained CLIP model~\cite{clip} as the image observation and language instruction encoder, and learn a manipulation policy on top for tasks across different object arrangements. While encouraging results have been shown in this line of research on generalization across training tasks, it is still very challenging for the learned policy to generalize to unseen tasks and environments. For example, as shown in Figure~\ref{fig:tasks} (a), our experiments show that if we train such a policy on two tasks including (i) pack objects of different shapes in the brown box and (ii) put blocks of different colors in the bowls of different colors, it is very challenging for the policy to generalize to a task that put objects of different shapes in different colored bowls. Furthermore, the difficulty drastically increases when we need to perform this task in the real world with a different robot. 

Our key insight is that, while the action generator (i.e., the policy) cannot generalize well, the action classifier with foundation models can still achieve a high accuracy even when ``zero-shot'' transferred to unseen environments~\cite{cliport,calvin}. In this paper, we leverage vision-language foundation models to provide feedback during deploying the policy in unseen tasks and environments. We utilize the feedback from foundation models to fine-tune the policy following test-time training~\cite{sun2020test, wangtent, kumar2021rma}, which updates model parameters during test-time. Specifically, we propose Policy Adaptation from Foundation model Feedback (PAFF) with a play and relabel pipeline. When adapting a trained policy to a new task or new environment, we first let the policy \textbf{play}, that is, the model continuously generates and performs actions given a series of language instructions in the new task and we record the demonstrations including the visual observations and model's actions. Of course, the instructions and the outcome demonstrations will often not match under the out-of-distribution environment. We then let the model \textbf{relabel} to make the correction, that is, the recorded demonstrations can be automatically relabeled by the vision-language pre-trained model. By taking the visual observations of recorded demonstrations as inputs, the pre-trained model can retrieve accurate language instructions correspondingly. Given the accurate paired demonstrations and instructions in the new environment, we can fine-tune and adapt the policy with them. We emphasize that the whole process of PAFF performs in an automatic way using trained models without human interventions. 

We carefully design a broad range of language conditioned robotic adaptation experiments to evaluate the policy adaptation across object composition, tasks and environments including from simulation to the real world. Our evaluations consist of (i) \textbf{Compositional Generalization} in Fig.~\ref{fig:tasks} (a), where we train a policy to pack objects of different shapes in the brown box, and put blocks of different colors in the bowls of different colors, and adapt it to put objects of different shapes in the bowls of different colors. (ii) \textbf{Out-of-distribution Generalization} in Fig.~\ref{fig:tasks} (c), where we train a policy to pack certain objects in the brown box, and adapt it to unseen objects; and in Fig.~\ref{fig:tasks} (d), where we adapt a policy trained on seen environments to an unseen environment with different textures and placements of static elements such as the sliding door, the drawer and the switch. (iii) \textbf{Sim-to-real Transfer} in Fig.~\ref{fig:tasks} (b), where we adapt a policy trained on simulation data to the real-world. 
We show PAFF improves baselines by a large-margin in all evaluations. In sim-to-real transfer, our method significantly improves the success rate by an average of 49.6\% on four tasks than the baseline. Our pipeline fills the domain gap between simulation and real world through utilizing the generalization capability of the foundation model. Our method also increases the success rate from 17.8\% to 35.0\% in the compositional generalization evaluation, and from 48.4\% to 63.8\% for packing unseen objects. When adapting the policy to an unseen environment, our method increases the success rate of completing 5 chains of language instructions from 5\% to 11\% over the baseline method. The extensive evaluation results show that PAFF can effectively adapt a language conditioned policy to unseen objects, tasks, environments, and realize sim-to-real transfer.

\begin{figure*}
	\centering
	\includegraphics[width=0.9\linewidth]{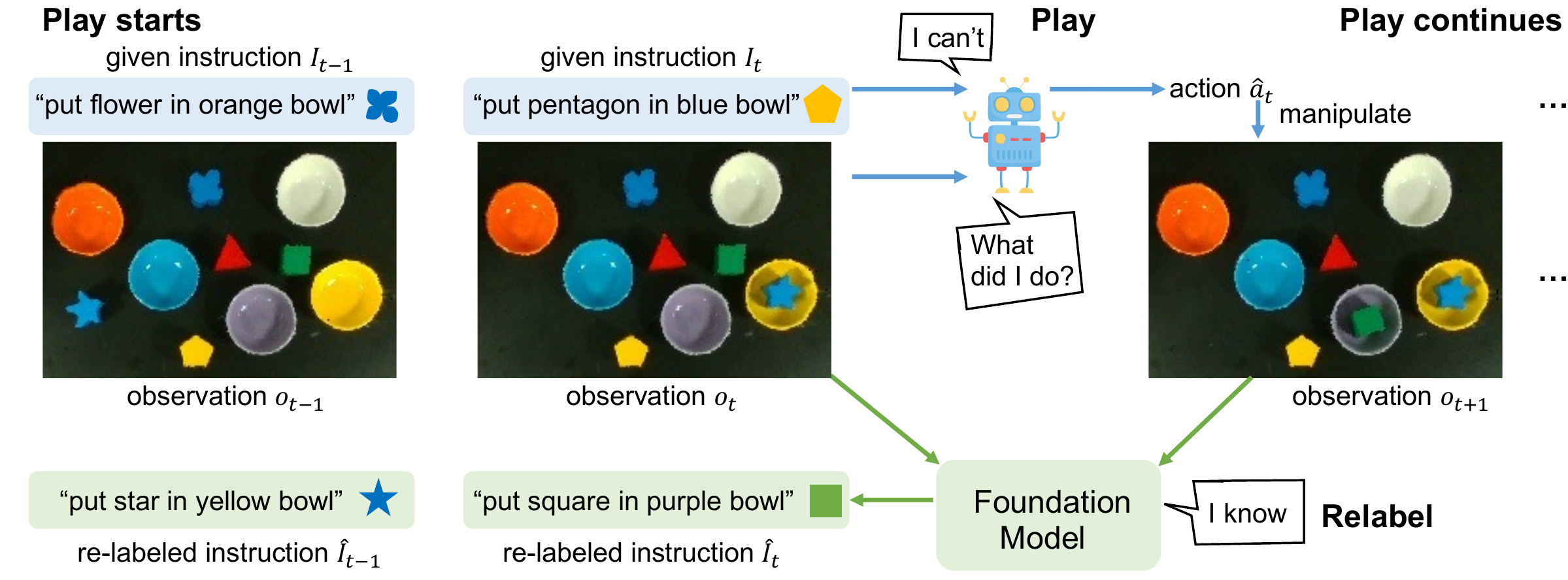}
  	\vspace{-5pt}	
	\caption{The pipeline of policy adaptation from foundation model feedback (PAFF). When we adapt a trained policy to a new task, we first let the robot \textbf{play}, that is, the policy continuously predicts and performs actions given a series of randomly generated language instructions. We record these demonstrations including visual observations and model's actions. After that, we let the model \textbf{relabel}, that is, the vision-language foundation model relabels the demonstrations by retrieving the language instructions given the recorded visual observations. We then fine-tune the policy with the paired observations and instructions, and corresponding actions, which are collected in an automatic way.}
	\vspace{-4mm}	
	\label{fig:method}
\end{figure*}

\section{Related Work}
\label{sec:related}
\textbf{Language Conditioned Manipulation.} Instruction-based manipulation has been a popular research topic in robotics~\cite{lang1,lang2,lang3,lang4,lang5,lang6,refer,grasp,play,calvin} not only because it provides an user-friendly interface, but also because the compositional properties of language allows skill generalization and guides long-horizon planning. Recently, the advancement of foundation models~\cite{clip,moco,instruct,devlin2018bert,raffel2020exploring,brown2020language,rae2021scaling,thoppilan2022lamda,wei2021finetuned} have led to significant progresses in generalizable manipulation skill learning~\cite{cliport,hulc,bc-zero,saycan2022arxiv}. These approaches adopt the pre-trained foundation models to encode the language instruction and visual observation, and train policy network on top. While this pipeline successfully generalizes the policy across different tasks and new objects, it still has a hard time on generalizing to completely unseen environments and tasks. In this paper, we propose policy adaptation from foundation model feedback to adapt our policy during deployment in unseen environments and tasks. 

\textbf{Policy Adaptation with Visual Inputs.} 
The ability to adapt a visuo-motor control policy to unseen environments is the key in many applications such as sim2real transfer. A popular way to achieve such generalization is using domain randomization~\cite{Tobin_2017,pinto2017asymmetric,Ramos_2019} and data augmentation~\cite{Lee2019ASR,cobbe2018quantifying,laskin2020reinforcement,zhang2020learning,hansen2021softda} for learning invariant visual representations. However, it is still very challenging for these policies to generalize  beyond the randomization range of training. Another line of research is to utilize GANs~\cite{cycle,goodfellow2020generative} to translate input images across domains before forwarding to the policy~\cite{gan1,gan2,zhang2019vr,zhang2020learning2}. However, all the approaches mentioned above require the tasks and environments are aligned across training and deployment. To adapt to a different task, researchers have proposed different ways to finetune the policy during deploying to the new environment~\cite{finetune1,finetune2,zhou2019environment,hansen2021deployment,nagabandi2018learning, kumar2021rma}. For example, Julian et al.~\cite{finetune2} propose to collect 800 grasping attempts when transferring to a new robotic manipulation environment, and use the collected data to fine-tune the policy. However, this process requires significant human interventions. Hansen et al.~\cite{hansen2021deployment} follows test-time training~\cite{sun2020test, wangtent} to allow the policy to continue training after deployment in a new environment by exploring self-supervision. Instead of using self-supervision, we utilize the feedback from vision-language foundation models to fine-tuning the policy in an automatic way. 

\textbf{Pre-trained Foundation Models in Robotics.} Besides language-conditioned manipulation tasks, we have witnessed the utilization of foundation models on various robotic learning algorithms~\cite{pretrain1, pretrain2, pretrain3,parisi2022unsurprising,seo2022reinforcement,gupta2022maskvit,ze2022visual,xiao2022robotic,agarwalreincarnating}. For example, Radosavovic et al.~\cite{pretrain1} propose to train a masked autoencoder~\cite{he2022masked} on large scale data such as  Ego4D~\cite{ego4d}, and perform real robot learning with behavior cloning using this pre-trained representation. It is shown that such pre-training helps largely reduce the sample efficiency and improves the performance. However, it is not thoroughly studied how pre-training can affect out-of-distribution generalization, and most works only use the pre-trained network as policy encoder. In this work, besides utilizing the representations from foundation models, we also exploit their generalization ability in recognizing visual concepts when transferred to unseen environments to provide feedback during deployment in a new environment. 

\section{Method}
We propose policy adaptation from foundation model feedback (PAFF) with a play and relabel pipeline. As summarized in Algorithm~\ref{alg:train}, our pipeline consists of two stages. In the first stage, we train a policy and fine-tune a vision-language pre-trained model with the training demonstrations. The second stage involves play and relabel in the new task. Specifically, as shown in Fig.~\ref{fig:method}, we first let the policy \textbf{play} with a series of randomly generated instructions and record the demonstrations including visual observations and actions. While the manipulation could be wrong, we let the fine-tuned vision-language model \textbf{relabel} the demonstrations by retrieving the language instructions given the visual observations. With the re-labeled pairs of demonstration-instruction data, we fine-tune the policy to adapt to the new task.

Our method utilizes the generalization capability of the vision-language foundation model in recognizing visual concepts, to annotate the data for fine-tuning the policy of the new task in an automatic way. PAFF can effectively adapt a language conditioned policy to manipulate unseen objects and solve new tasks in novel environments, and realize sim-to-real transfer.

\begin{algorithm}[h] 
\caption{PAFF} 
\label{alg:train} 
\begin{algorithmic}[1] 
    \Ensure
    train a policy and fine-tune a foundation model
    \Statex \Comment{\textbf{Stage 1}, Sec.~\ref{sec:3.1} and Sec.~\ref{sec:3.3}}
    \Require 
    training demonstrations; a randomly initialized policy ${\pi}_{\theta}$; a pre-trained foundation model ${f}_{\phi}$
    \For{each training demonstration}
    \State optimize the policy with $\mathcal{L}({\pi}_{\theta}(o_t,l_t),a_t)$
       \State optimize the pre-trained foundation model with $\mathcal{L}({f}_{\phi}(o_t,o_{t+1},l_t))$ 
        \EndFor
        \Statex \quad
        \Ensure
        play and relabel in a new task, then fine-tune the policy \Comment{\textbf{Stage 2}, Sec.~\ref{sec:3.2}}
        \Require
        initial observation; a series of language instructions
        \For{each language instruction}
        \State predict action $\hat{a}_t={\pi}_{\theta}(o_t,l_t)$, perform manipulation and record demonstrations
         \EndFor
         \For{each recorded demonstration}
         \State retrieve a language instruction $\hat{l}_t={f}_{\phi}(o_t,o_{t+1})$
         \EndFor
          \For{each recorded demonstration}
          \State optimize the policy with $\mathcal{L}({\pi}_{\theta}(o_t,\hat{l}_t),\hat{a}_t)$ 
          \EndFor
\end{algorithmic} 
\end{algorithm}

\subsection{Language Conditioned Policy}
\label{sec:3.1}
Inspired by previous work~\cite{cliport,hulc}, we consider the problem of learning a language conditioned policy $\pi$ that outputs actions $a_t$ given input ${\gamma}_t=(o_t,I_t)$ consisting of a visual observation $o_t$ and a language instruction $I_t$ as below:
\begin{equation}
\pi({\gamma}_t)=\pi(o_t,I_t) \rightarrow a_t
\end{equation}
We use an imitation-learning based method to learn the language conditioned policy. We experiment with two types of manipulation platforms with different action space: (i) We follow  CLIPORT~\cite{cliport} to formulate tabletop object manipulation (\eg, pick up a block and place it in a bowl) as a series of pick-and-place affordance predictions, where the objective is to detect actions and each action involves a start and final end-effector pose, and build our model upon a two-stream architecture in CLIPORT; (ii) We follow CALVIN~\cite{calvin} to formulate manipulation tasks that require continuous control (\eg, push the sliding door to the left side) as 7-DoF control, and build our model based on a hierarchical architecture in HULC~\cite{hulc}. We adopt the same imitation training loss $\mathcal{L}({\pi}_{\theta}(o_t,l_t),a_t)$ as defined in CLIPORT and HULC to optimize the policy.

\subsection{Play and Relabel}
\label{sec:3.2}
After we train a policy and adapt it to a new task, it can often make mistakes (\eg the policy picks up a star and places it in a yellow bowl given the instruction ``put the flower in the orange bowl'' as shown in Fig.~\ref{fig:method}). If our model can correct the instruction to a matching one (\eg ``put the star in the yellow bowl''), then the paired demonstration-instruction data can be used to fine-tune the policy.

We propose policy adaptation from foundation model feedback to collect the data for fine-tuning the policy of a new task in an automatic way, without human interventions. Specifically, as shown in Fig.~\ref{fig:method}, we first let the policy ``play'' with a series of randomly generated language instructions. Given the current visual observation $o_t$ and an instruction $I_t$, the policy predicts the action $a_t$ and the robot performs the corresponding manipulation, reaching a new scene with the visual observation $o_{t+1}$. After a certain number of language instructions are executed, the scene will be reset automatically by the robot, which moves the objects out of the containers to the table following the instruction ``move the objects out''. In this way, the policy can ``play'' in the new task, that is, the robot will explore the scene by continuously receiving language instructions and manipulating objects. We record these demonstrations including the visual observations $\{o_t\}_{t=1}^{T}$ and the robot's actions $\{\hat{a}_t\}_{t=1}^{T}$ predicted by the trained policy. 

After recording the demonstrations, we let the model relabel the demonstrations using the fine-tuned vision-language foundation model as shown in Fig.~\ref{fig:method}. We formulate the task of labeling the demonstrations as visual-to-language retrieval. Specifically, given the visual observations $o_t$ and $o_{t+1}$, the foundation model retrieves a language instruction $\hat{I}_t$ among all possible language instructions. Benefit from pre-training on large-scale data, the foundation model can generalize well across domains, thus is able to retrieve accurate language instructions for the recorded demonstrations. 

After play and relabel, we collect new pairs of demonstration-instruction data automatically including the visual observations $\{o_t\}_{t=1}^{T}$, the re-labeled language instructions $\{\hat{I}_t\}_{t=1}^{T}$, and the actions $\{\hat{a}_t\}_{t=1}^{T}$, which are used to fine-tune the policy for the adaptation to the new task.

\begin{figure}
	\centering
	\includegraphics[width=1.0\linewidth]{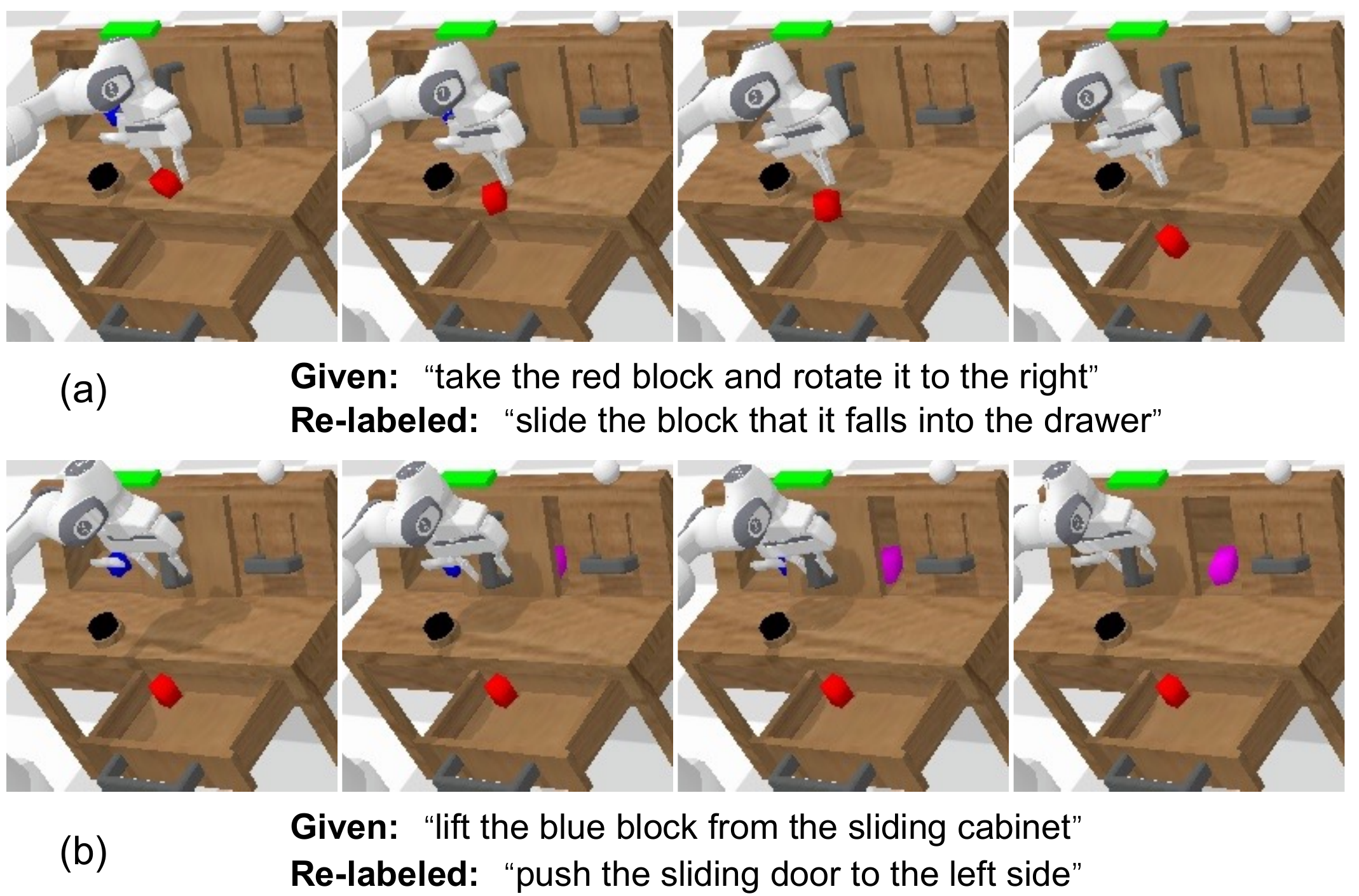}
  	\vspace{-15pt}	
	\caption{Given the language instruction in an unseen environment, the policy performs actions that do not match the instruction. The vision-language foundation model relabels the demonstration by taking the sequential visual observations as inputs and retrieving the language instruction among all possible instructions.}
	\vspace{-10pt}	
	\label{fig:relabel}
\end{figure}
 
\subsection{Vision-language Foundation Model}
\label{sec:3.3}
To make the model relabel the recorded demonstrations with the accurate instruction automatically, we cast the task of labeling demonstrations as visual-to-language retrieval, which is a mainstream downstream task in vision-language pre-training. The existing vision-language foundation model such as CLIP~\cite{clip} pre-trains on single images and their captions describing the static content, thus can not be directly applied to label the recorded demonstrations with sequential visual observations. For example, for manipulations that require continuous control such as ``pushing the sliding door to the left side'' as shown in Fig.~\ref{fig:relabel}, in order to recognize this action, the model requires the capability of both spatial and temporal reasoning among the sequential visual observations. The demonstration can not be labeled if the model lacks the ability to infer temporal information of the visual observations.

We utilize a Spatio-Temporal Adapter (ST-Adapter)~\cite{adapter} to fine-tune CLIP for the ability to reason about sequential visual observations of the recorded demonstrations. Specifically, the pre-trained CLIP $f_{\phi}$ adopts a visual encoder $f_v$ for visual representations, and a text encoder $f_t$ for language representations. We add a depth-wise 3D convolution layer~\cite{x3d} between the transformer~\cite{transformer} layers of the CLIP visual encoder $f_v$, so that $f_v$ can take sequential visual observations $o_t$ and $o_{t+1}$ as inputs to conduct spatio-temporal modeling (We denote sequential observations of executing a language instruction as $o_t$ and $o_{t+1}$ for simplicity). We use the CLIP text encoder $f_t$ to extract representations of the language instruction $l_t$. We adopt contrastive learning to maximize the similarity between the representations of visual observations denoted as $f_v(o_t,o_{t+1})$ and representations of the corresponding language instruction denoted as $f_t(l_t)$, and minimize the similarity between $f_v(o_t,o_{t+1})$ and representations of other language instructions with Noise-Contrastive Estimation (NCE)~\cite{oord2018representation},
\begin{small}
\begin{equation}
\mathcal{L}(f_{\phi}(o_t,o_{t+1},l_t)) = \text{NCE}(f_v(o_t,o_{t+1}),f_t(l_t)) 
\end{equation}
\end{small}
\begin{small}
    \begin{equation}
        \text{NCE}(x_i,y_i)= - \text{log}\frac{\text{exp}(x_i^Ty_i/\tau)}{\sum_{j=1}^N\text{exp}(x_i^Ty_j/\tau)} 
    \end{equation}
    \end{small}
where $N$ is the number of the language instructions and $\tau$ is the temperature hyper-parameter and we set $\tau=0.05$. We train the 3D convolution layers on the the seen tasks and freeze the parameters of the pre-trained CLIP. 

After fine-tuning the vision-language foundation model, given a recorded demonstration, we calculate the similarity between the representations of the visual observations and the representation of all language instructions, and retrieve the language instruction with the maximum similarity. The fine-tuned foundation model can achieve a high retrieval accuracy even when ``zero-shot'' transferred to unseen environments. For example, our fine-tuned CLIP achieves a retrieval accuracy of 99.3\% on the demonstrations of the unseen environment in the CALVIN~\cite{calvin} benchmark. Through utilizing the generalization capability of the large-scale vision-language foundation model in recognizing visual concepts, our method can relabel the recorded demonstrations for fine-tuning the policy in an automatic way.

\section{Experiment}
\label{sec:exp}
\subsection{Evaluation Settings}
We carefully design four language conditioned policy adaptation evaluations based on CLIPORT~\cite{cliport} and CALVIN~\cite{calvin}, to evaluate (i) \textbf{Compositional Generalization}, where we train a policy to pack objects of different shapes in the brown box (``pack-shapes''), and put blocks of different colors in the bowls of different colors (``put-blocks-in-bowls''), and adapt it to put objects of different shapes in the bowls of different colors (``put-shapes-in-bowls'') as shown in Fig.~\ref{fig:tasks} (a); (ii) \textbf{Out-of-distribution Generalization}, where we train a policy on packing seen objects and adapt it to unseen objects (``pack-unseen-objects'') using Google Scanned Objects dataset~\cite{downs2022google} as shown in Fig.~\ref{fig:tasks} (c), with the same split as CLIPORT, and train a policy on seen environments and adapt it to a new environment with different textures and differently positioned static elements such as the sliding door and light button in CALVIN as shown in Fig.~\ref{fig:tasks} (d). We use environment A, B and C for training, and environment D for adaptation; (iii) \textbf{Sim-to-real Transfer}, where we train a policy on simulation data and adapt it to real world with four tasks including ``pack-blocks'', ``packing-shapes'', ``put-blocks-in-bowls'', ``put-shapes-in-bowls'' as shown in Fig.~\ref{fig:tasks} (b). We also explore a challenging task (\ie compositional ``put-shapes-in-bowls''), where the policy is trained on simulation data of ``packing-shapes'' and ``put-blocks-in-bowls''.

For the compositional and out-of-distribution evaluations in the CLIPORT platform, we follow CLIPORT to report task success rate of 100 evaluation instances on 10 different scenes (with different blocks, objects and bowls), where the success rate is the number of the correctly placed objects, divided by the total number of the objects. For the out-of-distribution evaluation in the CALVIN platform, we follow CALVIN to evaluate Long-Horizon Multi-Task Language Control (LH-MTLC), which treats the 34 tasks as subtasks and evaluates 100 unique instruction chains, each consisting of five sequential tasks. The policy receives the next subtask in a chain only if it successfully completes the current one. We calculate the success rate of each task in the chain and the averaged successful sequence length as the evaluation metrics. For the sim-to-real transfer evaluation, we report task success rate of 10 evaluation instances for each task, where a instance consists of executing 5 language instructions.

\begin{figure*}
	\centering
	\includegraphics[width=1.0\linewidth]{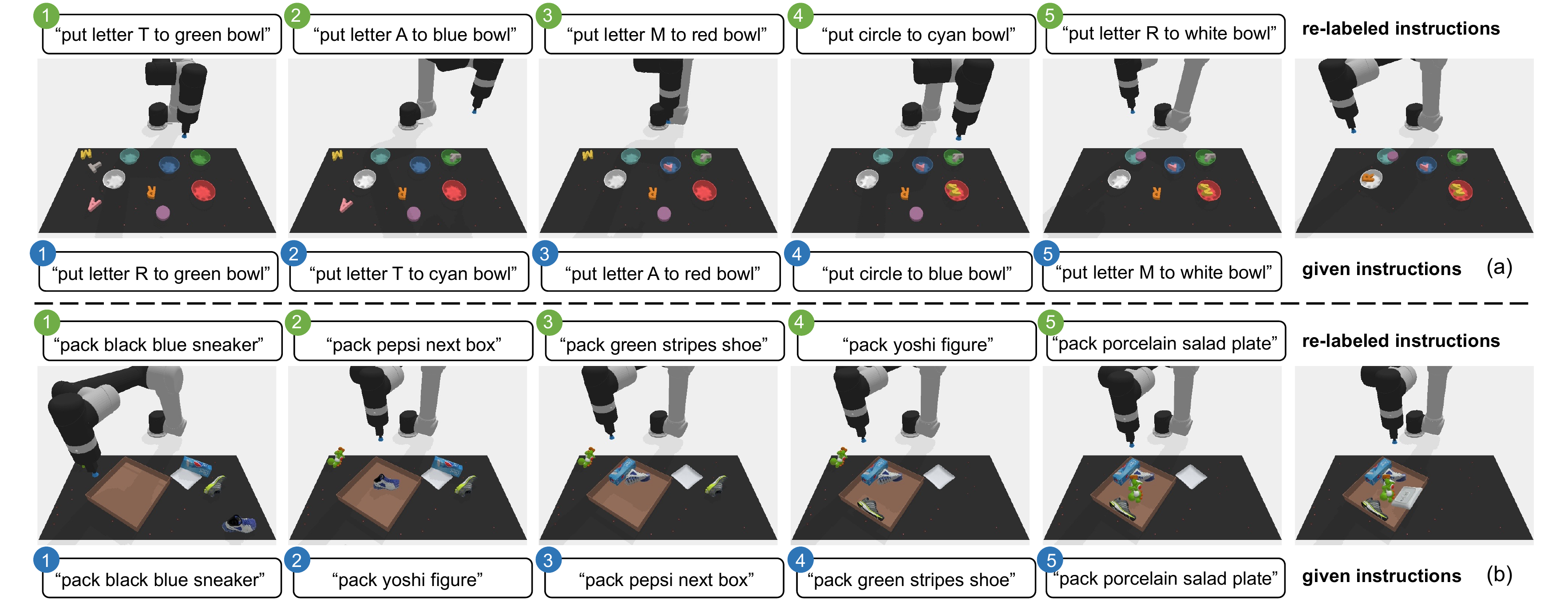}
  	\vspace{-15pt}	
	\caption{Given instructions vs. automatically re-labeled instructions of the recorded demonstrations with five consecutive steps, where we let the policy play in new tasks. The policy performs manipulations, where the outcome demonstrations do not match the instructions. By taking the recorded demonstrations as inputs, our model retrieves accurate language instructions correspondingly.}
	\vspace{-5pt}	
	\label{fig:relabel-cliport}
\end{figure*}

\begin{table}\centering\small
	\vspace{-5pt}
	\caption{Results of compositional and out-of-distribution generalization evaluation in the CLIPORT platform. The evaluation metric is the success rate, where each step receives a new instruction in the left column, while a step receives a new instruction when the previous instruction is executed correctly in the right column.} 
	\vspace{-5pt} 
	\scalebox{0.95}{
		\begin{tabular}{c|p{1cm}<{\centering}|p{1cm}<{\centering}|p{1cm}<{\centering}|p{1cm}<{\centering}}
			\toprule[1pt]
			Method & \multicolumn{2}{c|}{put-shapes-in-bowls} & \multicolumn{2}{c}{pack-unseen-objects}\\
			\hline
			CLIPORT~\cite{cliport}&28.0\%&16.8\%&58.9\%&46.1\%\\
			MdetrORT~\cite{mdetr}&33.8\%&17.8\%&62.0\%&48.4\%\\
			AugORT~\cite{aug1}&34.4\%&18.9\%&63.1\%&49.0\%\\
			Ours&\textbf{51.0\%}&\textbf{35.0\%}&\textbf{72.8\%}&\textbf{63.8\%}\\
			\bottomrule[1pt]
	\end{tabular}}
	\vspace{-15pt}
	\label{tab:calvin}
\end{table}

\subsection{Implementation Details}
To train policies, we follow CLIPORT~\cite{cliport} and HULC~\cite{hulc}, but adopt the pre-trained MDETR~\cite{mdetr} as the visual and language encoder (See Sec.~\ref{sec:ablation} for details). To fine-tune the vision-language foundation model (\textit{i.e.}, CLIP~\cite{clip}) for labeling, in the CLIPORT platform, we use the start and end frame of the visual observations as the input of CLIP. In the CALVIN platform, a language instruction in the training set corresponds to sequential observations with 64 frames, and we sample 8 frames following TSN~\cite{wang2018temporal} as the input of CLIP. We freeze the pre-trained CLIP and train ST-Adapter~\cite{adapter} for 300 epochs with the training data of seen tasks and environments on 8 GPUs. 

For evaluations in the CLIPORT platform, we first train a policy with 100 demonstrations per task for 200 epochs on a single GPU. We then let the robot play by following language instructions generated by a template ``put \_ in \_''. We record 40 demonstrations continuously for each scene, where a demonstration performs five instructions. After the model relabels demonstrations, we save recorded demonstrations, where the similarity score between representations of visual observations and retrieved instruction is higher than 3.0. We fine-tune the policy on the saved demonstrations for 100 epochs.

For the evaluation in the CALVIN platform, we train a policy on the training data of environments A, B, C for 10 epochs on 16 GPUs.  We then let the robot play by following instructions that are randomly chosen from all instructions used in seen environments. We record 500 demonstrations in the environment D, where a demonstration performs five subtasks. We fine-tune the policy on the re-labeled and saved demonstrations for 5 epochs.

\subsection{Baselines}
Besides CLIPORT~\cite{cliport} and HULC~\cite{hulc}, we also adopt two methods MdetrORT and MdetrLC as the baselines by replacing the visual and language encoder in CLIPORT and HULC with the pre-trained MDETR~\cite{mdetr}. They train a policy of seen tasks in exactly the same way as our method without play and relabel. Based on MdetrORT and MdetrLC, we further adopt two methods AugORT and AugLC, which use data augmentation to train a policy following Pashevich et al.~\cite{aug1}.

\begin{table}\centering\small
	\vspace{-5pt}
	\caption{Results of out-of-distribution generalization evaluation in the CALVIN platform. The number denotes the success rate of each subtask in the chain, and ``Len'' denotes the averaged successful sequence length.} 
	\vspace{-5pt} 
	\scalebox{0.95}{
		\begin{tabular}{c|cccccc}
			\toprule[1pt]
			\multirow{1}*{Method}&1&2&3&4&5&Len\\
			\hline
			\multirow{1}*{HULC~\cite{hulc}}&43\%&14\%&4\%&1\%&0\%&0.62\\
			\multirow{1}*{MdetrLC~\cite{mdetr}}&69\%&38\%&20\%&7\%&4\%&1.38\\
			\multirow{1}*{AugLC~\cite{aug1}}&69\%&43\%&22\%&9\%&5\%&1.48\\
			\multirow{1}*{Ours}&\textbf{72\%}&\textbf{47\%}&\textbf{30\%}&\textbf{13\%}&\textbf{11\%}&\textbf{1.73}\\
			\bottomrule[1pt]
	\end{tabular}}
	\vspace{-10pt}
	\label{tab:cliport}
\end{table}

\begin{figure*}
	\centering
	\includegraphics[width=1.0\linewidth]{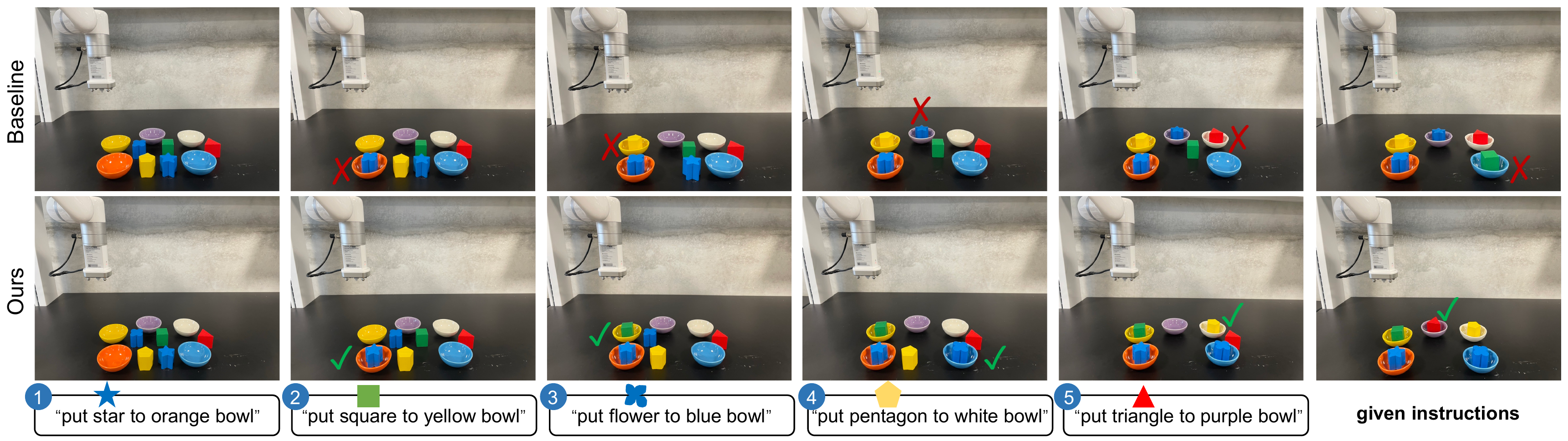}
 	\vspace{-15pt}	
	\caption{Real world experiments for sim-to-real transfer with five consecutive steps. The task puts objects of the specified shape in the bowls of the specified color.}
	\vspace{-5pt}	
	\label{fig:real}
\end{figure*}

\begin{table*}\centering\small
	
	\caption{Results of the sim-to-real transfer evaluation. The evaluation metric is the success rate, where each step receives a new language instruction. ``compositional'' denotes that the policy is trained on ``pack-shapes'' and ``put-blocks-in-bowls''.} 
	\vspace{-5pt} 
	\scalebox{1.0}{
		\begin{tabular}{c|c|c|c|c|c}
			\toprule[1pt]
			\multirow{1}*{Method}&pack-blocks&pack-shapes&put-blocks-in-bowls&put-shapes-in-bowls&put-shapes-in-bowls (compositional )\\
			\hline
			\multirow{1}*{MdetrORT~\cite{mdetr}}&72\%&32\%&40\%&24\%&14\%\\
			\multirow{1}*{Ours}&\textbf{98\%}&\textbf{92\%}&\textbf{88\%}&\textbf{82\%}&\textbf{70\%}\\
			\bottomrule[1pt]
	\end{tabular}}
	\vspace{-10pt}
	\label{tab:real}
\end{table*}

\subsection{Results}
\subsubsection{Compositional Generalization} 
As shown in Fig.~\ref{fig:relabel-cliport}, this kind of adaptation across object composition is challenging. For example, given the language instruction ``put letter T to cyan bowl'' in the second column, the trained policy wrongly picks up letter A object and puts it in the blue bowl in the third column. Such a mistake can be corrected by our fine-tuned vision-language foundation model as it retrieves ``put letter A to blue bowl'' as the language instruction given the visual observations of the second and the third column.

We show evaluation results of compositional generalization in Tab.~\ref{tab:cliport}. Our method achieves the best results under both evaluation protocols. Compared with MdetrORT~\cite{mdetr}, which trains a policy on seen tasks in exactly the same way as ours, our method improves the success rate by 17.2\%, which shows the effectiveness of our play and relabel pipeline for compositional generalization. The fine-tuned foundation model can be generalized to recognize the visual concepts (\ie objects and bowls) in the compositional setting as shown in Fig.~\ref{fig:relabel-cliport} (a), thus the re-labeled demonstrations can be used to fine-tune the policy in the new task. We further observe that MdetrORT performs better than CLIPORT by replacing the pre-trained CLIP~\cite{clip} with the pre-trained MDETR~\cite{mdetr} as the visual and language encoder. MDETR learns object-aware representations by aligning features of visual regions and text phrases, and can benefit the learning of a object manipulation policy. AugORT further improves performance slightly by applying data augmentation~\cite{aug1} during training, but still achieves poor results since compositional generalization is outside the randomization range of training. 

\subsubsection{Out-of-distribution Generalization}
We design two experiments to evaluate the out-of-distribution generalization. Tab.~\ref{tab:cliport} lists the results of packing unseen objects in the CLIPORT platform. Our method surpasses the baseline methods under both evaluation protocols. As shown in Fig.~\ref{fig:relabel-cliport}, when the policy picks up an object that does not match the language instruction, the vision-language foundation model can correct the mistake by retrieving a corresponding language instruction given the visual observations. Since the foundation model is pre-trained on large scale data, it can be generalized to recognizing unseen objects in the new task.

Tab.~\ref{tab:calvin} reports the results of adapting a policy to an unseen environment in the CALVIN platform.
Our method outperforms the baseline methods in terms of the success rate of each subtask in the chain and the averaged successful sequence length. As shown in Fig.~\ref{fig:relabel}, given the sequential visual observations, our fine-tuned vision-language foundation model is able to relabel the recorded demonstrations through spatial and temporal reasoning with the Spatio-Temporal Adapter~\cite{adapter}. We further observe that using the pre-trained MDETR as the visual and language encoder in HULC also improves performance, and applying the data augmentation helps adapt the policy to an unseen environment. But their results still lag far behind ours.

\subsubsection{Sim-to-real Transfer}
Adapting a policy trained on the simulation data to the real world is challenging because of the huge simulation-to-real gap such as different textures, lighting, colors, and objects. We evaluate four tasks in Tab.~\ref{tab:real}, and our method significantly boosts performance in each task compared with the baseline method (without play and relabel). We further evaluate a more difficult setting, which also involves the evaluation of compositional generalization by training policy on the simulation data of “pack-shapes” and “put-blocks-in-bowls''. As shown in Fig.~\ref{fig:real}, when we directly adapt the learned policy to the real world, it picks up the incorrect object and often places it in the bowl with wrong color. By contrast, through play and relabel, our method can perform manipulation correctly given the language instructions. Our pipeline fills the domain gap between simulation and real world through utilizing the generalization capability of the pre-trained foundation model.

\subsection{Ablation Study}
\label{sec:ablation}
{\flushleft \bf Number of recorded demonstrations.} We study the effect of the number of the recorded demonstrations when we let the policy ``play'' in the new task, where the policy executes five language instructions in each demonstration and then follows the instruction ``move the objects out'' to reset the scene automatically. As shown in Tab.~\ref{tab:number}, in general, recording 40 demonstrations for each scene achieves the best performance for both the compositional generalization and the out-of-distribution generalization evaluation. Recording more demonstrations increases the diversity of the collected data for fine-tuning the policy, thus benefits the policy adaptation to the new task.

\begin{table}\centering \small
	\vspace{0pt}
	\caption{Ablation Study on the number of the recorded demonstrations. Each demonstration consists of executing five language instructions. We evaluate the compositional generalization and the out-of-distribution generalization in the CLIPORT platform.} 
	\vspace{-5pt} 
	\scalebox{1.0}{
		\begin{tabular}{c|p{1cm}<{\centering}|p{1cm}<{\centering}|p{1cm}<{\centering}|p{1cm}<{\centering}}
			\toprule[1pt]
			Number & \multicolumn{2}{c|}{put-shapes-in-bowls} & \multicolumn{2}{c}{pack-unseen-objects}\\
			\hline
                10&36.0\%&21.3\%&69.7\%&59.0\%\\
			20&37.6\%&24.6\%&\textbf{73.2}\%&62.4\%\\
			40&\textbf{51.0\%}&\textbf{35.0\%}&72.8\%&\textbf{63.8\%}\\
			\bottomrule[1pt]
	\end{tabular}}
	\vspace{-15pt}
	\label{tab:number}
\end{table}

{\flushleft \bf Visual and language encoder} 
We explore different pre-trained foundation models including MoCo~\cite{moco}, DenseCL~\cite{densecl}, MAE~\cite{he2022masked}, GLIP~\cite{glip}, CLIP~\cite{clip} and MDETR~\cite{mdetr} as the visual and language encoder to train a policy, and evaluate their performances without play and relabel on the compositional generalization experiment with two evaluation protocols. For image pre-trained model, we use the text encoder in CLIP as the language encoder. The results are listed in Tab.~\ref{tab:encoder}. We have some observations. First of all, Vit~\cite{transformer} based MAE achieves the worst result. Compared with other models that use ResNet~\cite{resent} and Swin Transformer~\cite{swin} for multi-scale feature maps, Vit produces feature maps at a single scale, which does not benefit object manipulation policy learning. Second, DenseCL surpasses MoCo through performing dense pair-wise contrastive learning at the level of pixels rather than contrast global features. DenseCL tailors the self-supervised learning method for dense prediction tasks such as object detection, which is beneficial to learning a policy for object manipulation. Furthermore, GLIP and MDETR outperforms CLIP by aligning features of regions in the image and phrases in the text. The learned object-level and language-aware visual representations contributes to a better manipulation policy. Finally, MDETR achieves better results than GLIP, and we adopt this pre-trained foundation model as our visual and language encoder.

\begin{table}\centering\small
	\vspace{0pt}
	\caption{Ablation study on the visual and language encoder using different foundation models. We evaluate the compositional generalization with ``put-shapes-in-bowls'' in the CLIPORT platform.} 
	\vspace{-5pt} 
	\scalebox{1.0}{
		\begin{tabular}{c|c|c|c|c|c}
			\toprule[1pt]
        \multirow{1}*{MoCo}&DenseCL&MAE&GLIP&CLIP&MDETR\\
        \multirow{1}*{~\cite{moco}}&~\cite{densecl}&~\cite{he2022masked}&~\cite{glip}&~\cite{clip}&~\cite{mdetr}\\
			\hline
			22.6\%&26.4\%&19.0\%&30.1\%&28.0\%&\textbf{33.8\%}\\
                13.2\%&13.6\%&11.8\%&17.3\%&16.8\%&\textbf{17.8\%}\\
			\bottomrule[1pt]
	\end{tabular}}
	\vspace{-5pt}
	\label{tab:encoder}
\end{table}

\begin{table}\centering \small
	\vspace{0pt}
	\caption{Ablation study on the temporal reasoning mechanism for fine-tuning the foundation model. We evaluate the out-of-distribution generalization in the CALVIN platform with the averaged successful sequence length as the evaluation metric.} 
	\vspace{-5pt} 
	\scalebox{0.95}{
		\begin{tabular}{c|c|c|c}
			\toprule[1pt]
			2D joint-&3D joint-&Divided-&ST-Adapter\\           attention~\cite{time}&attention~\cite{videomae,kaiming_videomae}&attention~\cite{mcq}&\cite{adapter}\\
			\hline
                 1.60&1.53&1.46&\textbf{1.73}\\
			\bottomrule[1pt]
	\end{tabular}}
	\vspace{-10pt}
	\label{tab:temporal}
\end{table}

{\flushleft \bf Temporal reasoning  mechanism.} 
To capitalize the foundation model CLIP pre-trained on single images and captions for labeling recorded demonstrations with sequential visual observations, we explore different temporal reasoning mechanism to fine-tune CLIP including (i) 2d joint-attention~\cite{time}, which uses 2d convolution to flatten each image and performs joint-attention~\cite{transformer}, (ii) 3d joint-attention~\cite{videomae,kaiming_videomae}, which uses 3d convolution to flatten two images and performs joint-attention, (iii) divided-attention~\cite{mcq}, which adds temporal attention~\cite{time} among different images, and (iv) Spatio-Temporal Adapter (ST-Adapter)~\cite{adapter}, which adds a depth-wise 3D convolution layer~\cite{x3d} between each transformer layer. As listed in Tab.~\ref{tab:temporal}, ST-Adapter achieves the best result on the out-of-distribution generalization evaluation in the CALVIN platform. ST-Adapter trains the depth-wise 3D convolution layers for spatio-temporal reasoning, and meanwhile freezes the parameters of the pre-trained CLIP to reserve the generalization capability, thus can retrieve more accurate language instructions to fine-tune the policy.

\section{Conclusion}
In this work, we propose policy adaptation from foundation model feedback (PAFF), which leverages the vision-language foundation model to collect the data for fine-tuning the policy in unseen tasks and environments automatically. We evaluate our method on a broad range of language conditioned policy adaptation experiments including compositional generalization, out-of-distribution generalization and sim-to-real transfer, and show great superiority of our method. 
{\flushleft \bf \textbf{Acknowledgements.}} This project was supported, in part, by Amazon Research Award and gifts from Qualcomm.

{\small
\bibliographystyle{ieee_fullname}
\bibliography{egbib}
}

\end{document}